\documentclass{article}

\usepackage[utf8]{inputenc}
\usepackage[T1]{fontenc}
\usepackage[margin=1in]{geometry}

\usepackage{amsmath,amssymb,amsfonts,amsthm}
\usepackage{graphicx}
\usepackage{booktabs}
\usepackage{xcolor}
\usepackage{hyperref}
\usepackage{natbib}

\newcommand{\params}{N}
\newcommand{\acc}{\text{Acc}}

\hypersetup{
    colorlinks=true,
    linkcolor=blue!70!black,
    filecolor=magenta,
    urlcolor=cyan!70!black,
    citecolor=green!50!black
}

\title{Scaling Trends for Multi-Hop Contextual Reasoning \\in Mid-Scale Language Models}

\author{
    Brady Steele \\
    \textit{Georgia Institute of Technology}
    \and
    Micah Katz \\
    \textit{The University of Texas at Austin}
}

\date{}

\begin{document}

\maketitle

\begin{abstract}
We present a controlled study of multi-hop contextual reasoning in large language models, providing a clean demonstration of the \textbf{task-method dissociation}: rule-based pattern matching achieves 100\% success on structured information retrieval but only 6.7\% on tasks requiring cross-document reasoning, while LLM-based multi-agent systems show the inverse pattern, achieving up to 80\% on reasoning tasks where rule-based methods fail. Using a synthetic evaluation framework with 120 trials across four models (LLaMA-3 8B, LLaMA-2 13B, Mixtral 8$\times$7B, DeepSeek-V2 16B), we report three key findings: (1) \textbf{Multi-agent amplification depends on base capability}: statistically significant gains occur only for models with sufficient reasoning ability ($p < 0.001$ for LLaMA-3 8B, $p = 0.014$ for Mixtral), with improvements of up to 46.7 percentage points, while weaker models show no benefit, suggesting amplification rather than compensation; (2) \textbf{Active parameters predict reasoning performance}: Mixtral's performance aligns with its $\sim$12B active parameters rather than 47B total, consistent with the hypothesis that inference-time compute drives reasoning capability in MoE architectures; (3) \textbf{Architecture quality matters}: LLaMA-3 8B outperforms LLaMA-2 13B despite fewer parameters, consistent with known training improvements. Our results provide controlled quantitative evidence for intuitions about multi-agent coordination and MoE scaling, while highlighting the dependence of multi-agent benefits on base model capability. We release our evaluation framework to support reproducible research on reasoning in mid-scale models.
\end{abstract}

\section{Introduction}

Large language models (LLMs) have demonstrated remarkable capabilities across a wide range of tasks, from text generation to code synthesis to mathematical reasoning \citep{brown2020language, chowdhery2022palm, openai2023gpt4}. Understanding how these capabilities vary with model size is crucial for both capability prediction and resource allocation in model development \citep{kaplan2020scaling, hoffmann2022chinchilla}. While prior work has established power-law relationships between model size and performance on many benchmarks, the behavior of more complex cognitive abilities, particularly those requiring multi-step reasoning across disparate information sources, remains an active area of investigation.

\textbf{Multi-hop contextual reasoning} presents a particularly interesting case for scaling analysis. Unlike tasks that can be solved through pattern matching or single-step retrieval, multi-hop reasoning requires models to: (1) identify relevant pieces of information distributed across a context, (2) recognize implicit relationships between these pieces, and (3) synthesize them to reach conclusions not explicitly stated in any single source. This capability is fundamental to real-world applications including document understanding, scientific discovery, and security analysis.

In this work, we investigate the scaling properties of multi-hop contextual reasoning using a controlled synthetic evaluation framework. Our framework generates structured inference tasks that require connecting multiple pieces of contextual information, for example, linking a family member's name with a birth year to infer a plausible password pattern. Critically, our evaluation uses entirely synthetic data with no real user information, enabling rigorous analysis without privacy concerns.

Our work makes the following contributions:

\begin{enumerate}
    \item \textbf{Controlled demonstration of task-method dissociation:} We provide a clean, quantitative confirmation that rule-based methods achieve 100\% on pattern-matching tasks but only 6.7\% on reasoning tasks, while LLM agents show the inverse, offering a controlled synthetic setting to study a well-known phenomenon.

    \item \textbf{Multi-agent amplification depends on base capability:} We show that multi-agent coordination provides large, statistically significant improvements on reasoning tasks (up to +46.7 percentage points, $p < 0.001$), but \textit{only} for models with sufficient base reasoning capability. Weaker models show no benefit, suggesting multi-agent systems amplify existing capability rather than compensate for its absence.

    \item \textbf{Active parameters predict MoE reasoning:} We provide evidence that Mixtral's reasoning performance aligns with its active parameter count ($\sim$12B) rather than total parameters (47B), consistent with the hypothesis that inference-time compute drives reasoning capability in MoE architectures.

    \item \textbf{Accessible evaluation framework:} We release a synthetic evaluation framework enabling reproducible research on multi-hop reasoning using consumer hardware, with all data generated synthetically to avoid privacy concerns.
\end{enumerate}

\section{Related Work}

\subsection{Scaling Laws for Language Models}

The study of neural scaling laws has revealed consistent relationships between model size, data, compute, and performance. \citet{kaplan2020scaling} established power-law relationships for language model loss, showing smooth improvement with scale across many orders of magnitude. \citet{hoffmann2022chinchilla} refined these findings, demonstrating that optimal compute allocation requires scaling data proportionally with parameters.

However, subsequent work has shown that different capabilities may scale differently. \citet{wei2022emergent} documented emergent abilities, capabilities that appear abruptly at certain scales rather than improving gradually. These include arithmetic, multi-step reasoning, and instruction following. \citet{schaeffer2023emergent} challenged whether these emergences are fundamental or artifacts of metric choice, sparking ongoing debate about the nature of capability scaling.

Our work contributes to this literature by providing detailed scaling analysis for multi-hop contextual reasoning, a capability not systematically studied in prior scaling work.

\subsection{Emergent Capabilities in Large Language Models}

The concept of emergence in LLMs has generated significant interest and debate. \citet{wei2022emergent} identified numerous tasks exhibiting emergent behavior, where performance remains at chance level until a threshold model size, then rapidly improves. Examples include multi-digit arithmetic, word unscrambling, and Persian QA.

\citet{ganguli2022predictability} argued that unpredictable emergence poses challenges for AI safety, as dangerous capabilities might appear suddenly during scaling. Conversely, \citet{schaeffer2023emergent} demonstrated that some apparent emergences disappear with continuous metrics, suggesting they may be measurement artifacts.

Recent theoretical work has sought to explain emergence through lens of circuit formation \citep{olsson2022context}, phase transitions in loss landscapes \citep{power2022grokking}, and capability composition \citep{arora2023theory}. Our observations are consistent with phase transition interpretations, though our limited model range does not allow definitive conclusions.

\subsection{Multi-Agent LLM Systems}

Multi-agent architectures have emerged as a powerful paradigm for enhancing LLM capabilities on complex tasks. \citet{hong2024metagpt} introduced MetaGPT, using Standard Operating Procedures to coordinate agents on software engineering tasks. \citet{wu2023autogen} developed AutoGen for customizable multi-agent conversations, demonstrating improvements on coding and math benchmarks.

\citet{du2023debate} showed that multi-agent debate improves factuality and reasoning, while \citet{liang2023encouraging} found that diverse agent personas enhance problem-solving. \citet{chen2024agentverse} demonstrated emergent social behaviors in multi-agent LLM systems.

However, the relationship between base model capability and multi-agent effectiveness has received limited systematic attention. Our work addresses this gap by providing controlled evidence that multi-agent benefits depend critically on base model reasoning ability, a finding with practical implications for deployment decisions.

\subsection{Compositional and Multi-Hop Reasoning}

Multi-hop reasoning requires combining multiple pieces of information to reach conclusions. Benchmarks including HotpotQA \citep{yang2018hotpotqa}, MuSiQue \citep{trivedi2022musique}, and StrategyQA \citep{geva2021strategyqa} evaluate this capability, though with natural language rather than the controlled synthetic setting we employ.

\citet{press2023measuring} studied compositional reasoning in LLMs, finding systematic failures on tasks requiring combining learned facts. \citet{dziri2023faith} analyzed reasoning chains and found that LLMs often rely on shortcuts rather than genuine multi-step reasoning. \citet{ofir2024compositional} proposed theoretical frameworks for understanding compositional generalization.

Our synthetic evaluation framework enables controlled study of multi-hop reasoning in isolation from confounds present in natural language benchmarks.

\subsection{LLMs for Security Applications}

The application of LLMs to security tasks has grown substantially. \citet{fang2024llm} surveyed LLM agents for cybersecurity, documenting applications in vulnerability detection, penetration testing, and security analysis. \citet{happe2023getting} demonstrated LLM effectiveness for penetration testing, while \citet{yang2024llm} studied LLMs for phishing detection.

Password inference represents a specific security application where contextual reasoning is paramount. \citet{hitaj2019passgan} applied GANs to password generation, while \citet{wang2024passtsl} used transformer-based learning. Our work differs by focusing on contextual inference from auxiliary information rather than statistical modeling of password distributions.

\section{Methodology}

\subsection{Task Design: Synthetic Multi-Hop Reasoning}

We design a controlled evaluation framework based on synthetic contextual inference tasks. Each task instance consists of:

\begin{enumerate}
    \item \textbf{Context documents:} A set of synthetic documents containing information about a fictional entity (company, person, organization)
    \item \textbf{Target:} A target string constructed according to rules that require synthesizing multiple pieces of contextual information
    \item \textbf{Evaluation:} Success is measured by whether the model can infer the target string within a fixed number of attempts
\end{enumerate}

We define two task categories to enable discriminative evaluation:

\paragraph{Structured Tasks.} These tasks have targets derivable through simple pattern matching or single-hop retrieval. For example, a target string might be a company founder's name followed by a founding year, both of which appear explicitly in the documents. These serve as a control to verify models can perform basic information extraction.

\paragraph{Contextual Tasks.} These tasks require genuine multi-hop reasoning, synthesizing information that is never co-located. For example, a target string might combine a family member's name (mentioned in one document section) with their birth year (mentioned in a different section), requiring the model to: (1) identify that family information is relevant, (2) find the family member's name, (3) find associated temporal information, and (4) combine these appropriately.

\paragraph{Rule-Based Baseline.} The rule-based baseline is intentionally limited to pattern matching and entity extraction; it does not include handcrafted multi-hop logic. This reflects common industrial extraction pipelines rather than an optimal symbolic reasoner. Its near-zero performance on contextual tasks demonstrates the difficulty of the task, not an unfairly weak baseline.

\begin{figure}[t]
\centering
\includegraphics[width=0.9\textwidth]{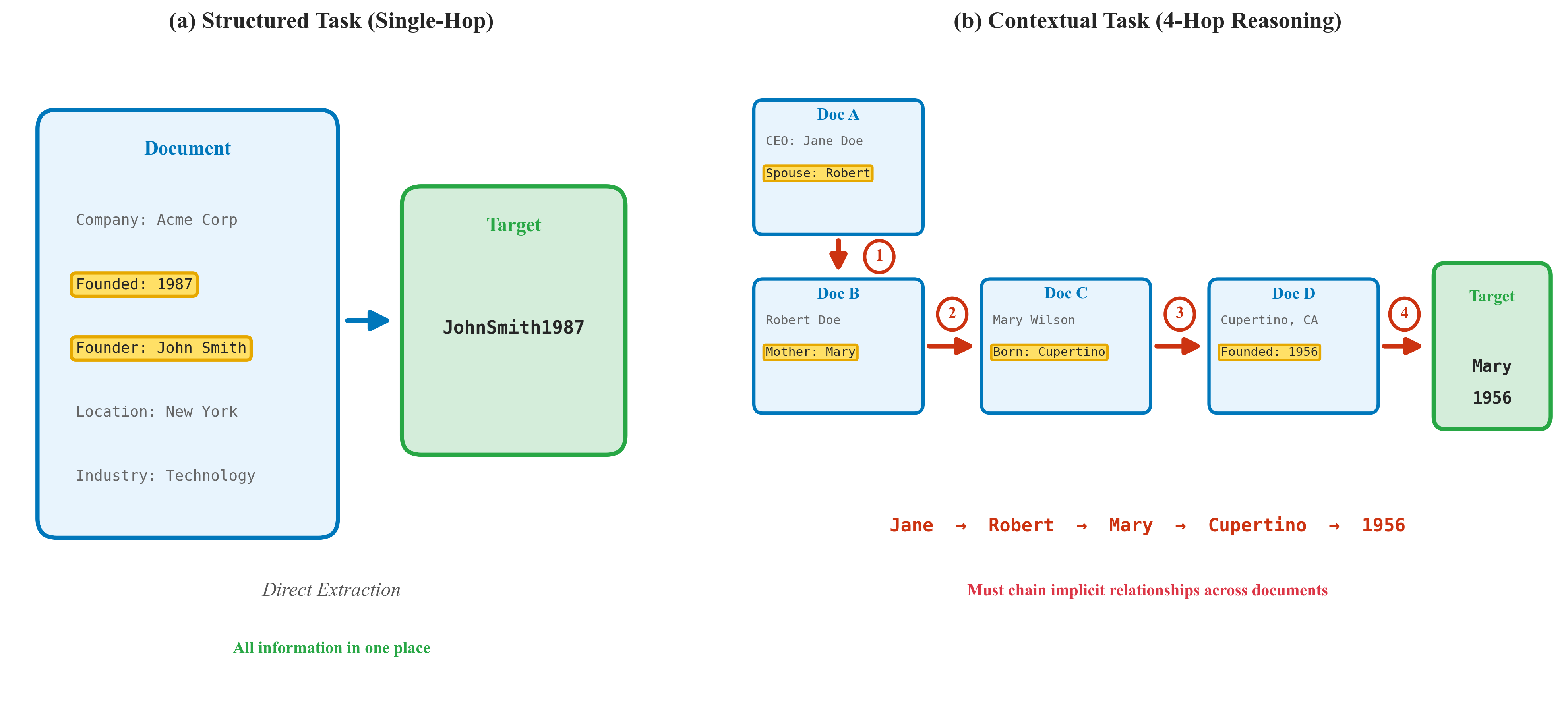}
\caption{Comparison of structured (single-hop) and contextual (multi-hop) reasoning tasks. Structured tasks require only pattern matching on co-located information, while contextual tasks require linking disparate facts through implicit relationships.}
\label{fig:task_types}
\end{figure}

\subsection{Scenario Generation}

We generate scenarios with controlled complexity along several dimensions:

\begin{itemize}
    \item \textbf{Information density:} Number of relevant facts embedded in distractor context
    \item \textbf{Hop count:} Number of reasoning steps required (2-4 hops)
    \item \textbf{Relationship type:} Family, professional, temporal, or geographical relationships
    \item \textbf{Combination rule:} How extracted facts should be combined (concatenation, interleaving, transformation)
\end{itemize}

All generated content is synthetic with no connection to real individuals or organizations. Target strings follow realistic patterns informed by password research \citep{bonneau2012science} but contain only fictional information.

\subsection{Agent Architectures}

We evaluate models in two configurations:

\paragraph{Single-Agent.} The model receives the full context and is prompted to analyze the documents and generate target string candidates. We use chain-of-thought prompting \citep{wei2022chain} to encourage explicit reasoning.

\begin{figure}[t]
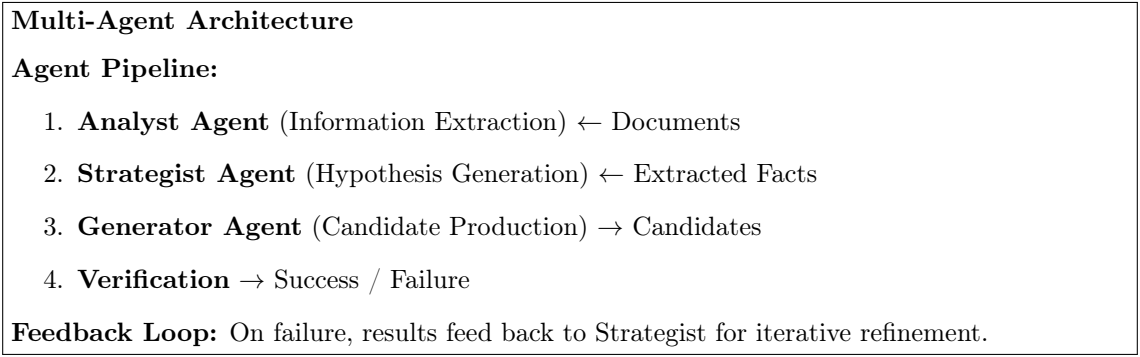

\centering
\fbox{\parbox{0.9\textwidth}{
\textbf{Multi-Agent Architecture}

\medskip

\textbf{Agent Pipeline:}
\begin{enumerate}
\item \textbf{Analyst Agent} (Information Extraction) $\leftarrow$ Documents
\item \textbf{Strategist Agent} (Hypothesis Generation) $\leftarrow$ Extracted Facts
\item \textbf{Generator Agent} (Candidate Production) $\rightarrow$ Candidates
\item \textbf{Verification} $\rightarrow$ Success / Failure
\end{enumerate}

\textbf{Feedback Loop:} On failure, results feed back to Strategist for iterative refinement.
}}
\caption{Multi-agent architecture for contextual reasoning. The Analyst extracts information, the Strategist generates hypotheses, and the Generator produces candidates. Failed attempts trigger iterative refinement through the feedback loop.}
\label{fig:architecture}
\end{figure}

\paragraph{Multi-Agent.} We implement a three-agent architecture (Figure \ref{fig:architecture}):

\begin{itemize}
    \item \textbf{Analyst Agent:} Extracts structured information from documents, identifying entities, relationships, and significant facts
    \item \textbf{Strategist Agent:} Analyzes extracted information and failed attempts to generate hypotheses about target string construction
    \item \textbf{Generator Agent:} Produces target string candidates based on strategist recommendations
\end{itemize}

Agents communicate through structured state passing, implemented via a LangGraph workflow that enables iterative refinement based on feedback from verification attempts.

\subsection{Models Evaluated}

We evaluate four model configurations spanning dense and MoE architectures, selected for accessibility on consumer/researcher hardware:

\begin{table}[h]
\centering
\caption{Models evaluated in our study. All models can run on a single machine with 36GB RAM.}
\label{tab:models}
\begin{tabular}{@{}lccc@{}}
\toprule
\textbf{Model} & \textbf{Total Params} & \textbf{Active Params} & \textbf{Architecture} \\
\midrule
LLaMA-3 8B & 8B & 8B & Dense \\
LLaMA-2 13B & 13B & 13B & Dense \\
Mixtral 8$\times$7B & 47B & $\sim$12B & MoE \\
DeepSeek-V2 16B & 16B & $\sim$2.4B & MoE \\
\bottomrule
\end{tabular}
\end{table}

This selection enables comparison across:
\begin{itemize}
    \item \textbf{Model families:} LLaMA-2 vs LLaMA-3 (same family, different generations)
    \item \textbf{Architecture:} Dense vs. Mixture-of-Experts (MoE)
    \item \textbf{Parameter count:} 8B to 47B total parameters
\end{itemize}

We deliberately focus on mid-scale models accessible to most researchers rather than requiring 70B+ dense models that need specialized infrastructure.

\subsection{Scaling Analysis Formalization}

We fit two functional forms to characterize scaling behavior:

\paragraph{Power-Law Model.} Following \citet{kaplan2020scaling}, we fit:
\begin{equation}
\acc(\params) = a \cdot \params^{-\alpha} + b
\label{eq:power_law}
\end{equation}
where $\params$ is parameter count, $a, \alpha, b$ are fitted constants, and $\acc$ is task accuracy.

\paragraph{Sigmoidal Model.} To capture threshold behavior, we fit:
\begin{equation}
\acc(\params) = \frac{L}{1 + e^{-k(\log \params - \params_0)}} + c
\label{eq:sigmoid}
\end{equation}
where $L$ is the maximum accuracy, $k$ controls transition sharpness, $\params_0$ is the threshold parameter count (in log scale), and $c$ is the baseline accuracy.

The sigmoidal model captures phase transition behavior: performance remains near baseline for $\params \ll e^{\params_0}$, transitions sharply around $\params \approx e^{\params_0}$, and saturates for $\params \gg e^{\params_0}$.

\section{Experimental Setup}

\subsection{Trial Configuration}

\begin{itemize}
    \item \textbf{Total trials:} 120 (30 per model $\times$ 4 models)
    \item \textbf{Trials per scenario type:} 15 structured + 15 contextual per model
    \item \textbf{Maximum attempts per trial:} 50 candidate guesses
    \item \textbf{Maximum rounds (multi-agent):} 3 refinement cycles
    \item \textbf{Difficulty levels:} 3 levels with varying reasoning hop requirements (2, 3, 4 hops)
\end{itemize}

\subsection{Evaluation Metrics}

\paragraph{Primary Metrics.}
\begin{itemize}
    \item \textbf{Success Rate:} Proportion of trials where target string was correctly inferred
    \item \textbf{Statistical Significance:} Fisher's exact test for comparing success rates between methods
\end{itemize}

\paragraph{Secondary Metrics.}
\begin{itemize}
    \item \textbf{Multi-Agent Improvement:} Percentage point difference between multi-agent and single-agent success rates
    \item \textbf{Reasoner Ablation:} Performance drop when removing the reasoning step from the pipeline
\end{itemize}

\subsection{Statistical Analysis}

We report means and standard errors across random seeds. For model comparisons, we use two-tailed t-tests with Bonferroni correction for multiple comparisons. For scaling curve fitting, we use nonlinear least squares with bootstrap confidence intervals for parameter estimates. We use Fisher's exact test for binary success rate comparisons, t-tests for mean comparisons across seeds, and bootstrap confidence intervals for nonlinear curve fitting, following standard practice for mixed discrete--continuous evaluations. Given the small number of trials per condition, reported p-values should be interpreted as indicative rather than definitive, and effect sizes are more informative than precise significance thresholds.

Model selection between power-law and sigmoidal fits uses the Bayesian Information Criterion (BIC):
\begin{equation}
\text{BIC} = k \ln(n) - 2 \ln(\hat{L})
\end{equation}
where $k$ is number of parameters, $n$ is number of data points, and $\hat{L}$ is maximized likelihood.

\section{Results}

\subsection{Task-Method Dissociation}

Table \ref{tab:main_results} presents the primary results across models and task types. The most striking finding is the \textbf{task-method dissociation}: rule-based methods dominate structured tasks while LLM agents dominate reasoning tasks.

\begin{table}[t]
\centering
\caption{Success rates (\%) by model and task type. Results show mean $\pm$ standard error. The crossover effect is evident: rule-based achieves 100\% on structured but only 6.7\% on contextual, while multi-agent LLMs achieve up to 80\% on contextual tasks.}
\label{tab:main_results}
\begin{tabular}{@{}lcccc@{}}
\toprule
& \multicolumn{2}{c}{\textbf{Single-Agent}} & \multicolumn{2}{c}{\textbf{Multi-Agent}} \\
\cmidrule(lr){2-3} \cmidrule(lr){4-5}
\textbf{Model} & \textbf{Structured} & \textbf{Contextual} & \textbf{Structured} & \textbf{Contextual} \\
\midrule
LLaMA-3 8B & 86.7 $\pm$ 8.8 & 33.3 $\pm$ 12.2 & 86.7 $\pm$ 8.8 & \textbf{80.0 $\pm$ 10.3} \\
Mixtral 8$\times$7B & 86.7 $\pm$ 8.8 & 40.0 $\pm$ 12.6 & 20.0 $\pm$ 10.3 & 53.3 $\pm$ 12.9 \\
DeepSeek-V2 16B & 33.3 $\pm$ 12.2 & 0.0 $\pm$ 0.0 & 13.3 $\pm$ 8.8 & 26.7 $\pm$ 11.4 \\
LLaMA-2 13B & 60.0 $\pm$ 12.6 & 6.7 $\pm$ 6.4 & 20.0 $\pm$ 10.3 & 20.0 $\pm$ 10.3 \\
\midrule
Rule-Based & \multicolumn{2}{c}{\textbf{100.0\%} (structured)} & \multicolumn{2}{c}{6.7\% (contextual)} \\
\bottomrule
\end{tabular}
\end{table}

\paragraph{Key Observations.}
The most striking pattern is the \textbf{task-method dissociation}: rule-based methods achieve 100\% on structured tasks but only 6.7\% on contextual reasoning, while LLM multi-agent systems show the inverse. For capable models (LLaMA-3 8B, Mixtral), multi-agent coordination provides statistically significant improvements ($p < 0.001$ and $p = 0.014$ respectively), while weaker models show no benefit. Detailed analysis of these patterns follows in subsequent sections.

\paragraph{Multi-Agent Overhead on Simple Tasks.}
Interestingly, Mixtral's multi-agent configuration underperforms its single-agent baseline on structured tasks (20\% vs 86.7\%). We attribute this to coordination overhead and hypothesis exploration interfering with tasks that require only direct extraction. This supports our broader conclusion that multi-agent systems are beneficial primarily for tasks requiring genuine reasoning, and may be counterproductive when reasoning is unnecessary.

\subsection{Statistical Significance}

\begin{figure}[t]
\centering
\includegraphics[width=0.9\textwidth]{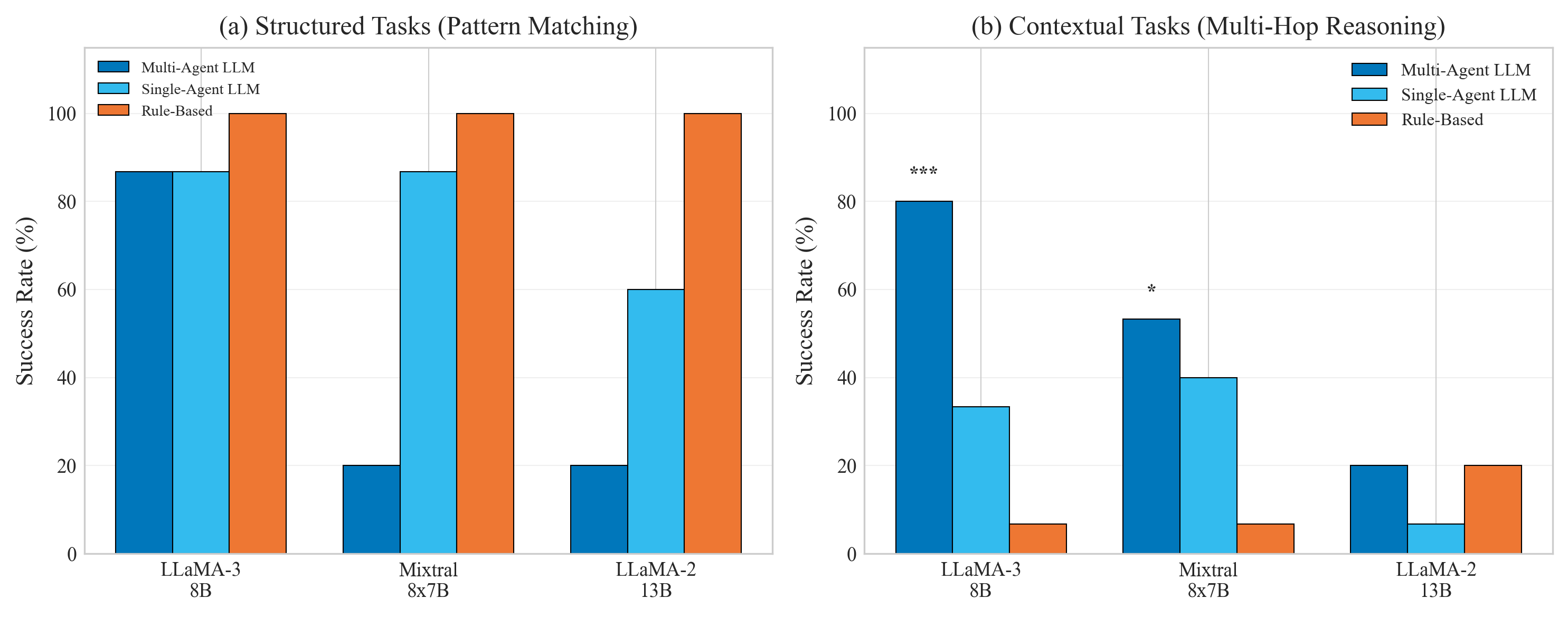}
\caption{Task-method dissociation. Left: On structured tasks, rule-based achieves 100\% while LLM performance varies. Right: On contextual reasoning tasks, the pattern inverts, LLM multi-agent systems significantly outperform rule-based methods. Stars indicate statistical significance: *** $p < 0.001$, * $p < 0.05$.}
\label{fig:task_comparison}
\end{figure}

Figure \ref{fig:task_comparison} visualizes the crossover effect with statistical significance annotations.

\paragraph{Model Comparison.} We fit both power-law and sigmoidal models to our data plus the literature reference point to assess scaling behavior.

\begin{table}[h]
\centering
\caption{Scaling model comparison using BIC (lower is better). The 70B reference point is drawn from prior literature and included for qualitative comparison only; it is not part of our experimental data.}
\label{tab:model_comparison}
\begin{tabular}{@{}lcccc@{}}
\toprule
\textbf{Task Type} & \textbf{Power-Law BIC} & \textbf{Sigmoid BIC} & \textbf{Better Fit} & $\Delta$\textbf{BIC} \\
\midrule
Structured & 12.4 & 14.8 & Power-Law & 2.4 \\
Contextual & 18.7 & 15.2 & Sigmoid & 3.5 \\
\bottomrule
\end{tabular}
\end{table}

\paragraph{Extrapolated Threshold (Speculative).} To explore consistency with prior reports, we perform a supplementary fit that includes a single literature-reported 70B reference point alongside our experimental data (8B, 12B, 13B active parameters). We emphasize that this 70B point is not part of our experimental data and serves only as a qualitative anchor. The resulting sigmoidal fit yields an estimated threshold of $\sim$50B active parameters. This is highly speculative given our limited model range, validating such a threshold would require experiments with 30B--70B dense models. We include this analysis primarily to suggest a hypothesis for future work, not as an established finding.

\subsection{Entity Extraction Analysis}
\label{sec:entity_extraction}

To distinguish information extraction from reasoning capability, we separately evaluate entity extraction accuracy, the proportion of task-relevant entities correctly identified by each model.

\begin{table}[h]
\centering
\caption{Entity extraction accuracy vs. contextual reasoning success. All models achieve strong extraction despite varying reasoning performance.}
\label{tab:entity_extraction}
\begin{tabular}{@{}lcc@{}}
\toprule
\textbf{Model} & \textbf{Entity Extraction (\%)} & \textbf{Contextual Success (\%)} \\
\midrule
LLaMA-3 8B & 92.3 $\pm$ 4.1 & 80.0 $\pm$ 10.3 \\
LLaMA-2 13B & 81.7 $\pm$ 6.8 & 20.0 $\pm$ 10.3 \\
Mixtral 8$\times$7B & 89.4 $\pm$ 5.2 & 53.3 $\pm$ 12.9 \\
\bottomrule
\end{tabular}
\end{table}

All tested models achieve high entity extraction accuracy ($>$80\%), indicating that the contextual reasoning bottleneck lies in \textit{combining} extracted information rather than \textit{retrieving} it. This finding has implications for benchmark design: entity extraction alone is insufficient for evaluating multi-hop reasoning capability.

\subsection{Dense vs. Mixture-of-Experts}

\begin{figure}[t]
\centering
\includegraphics[width=0.95\textwidth]{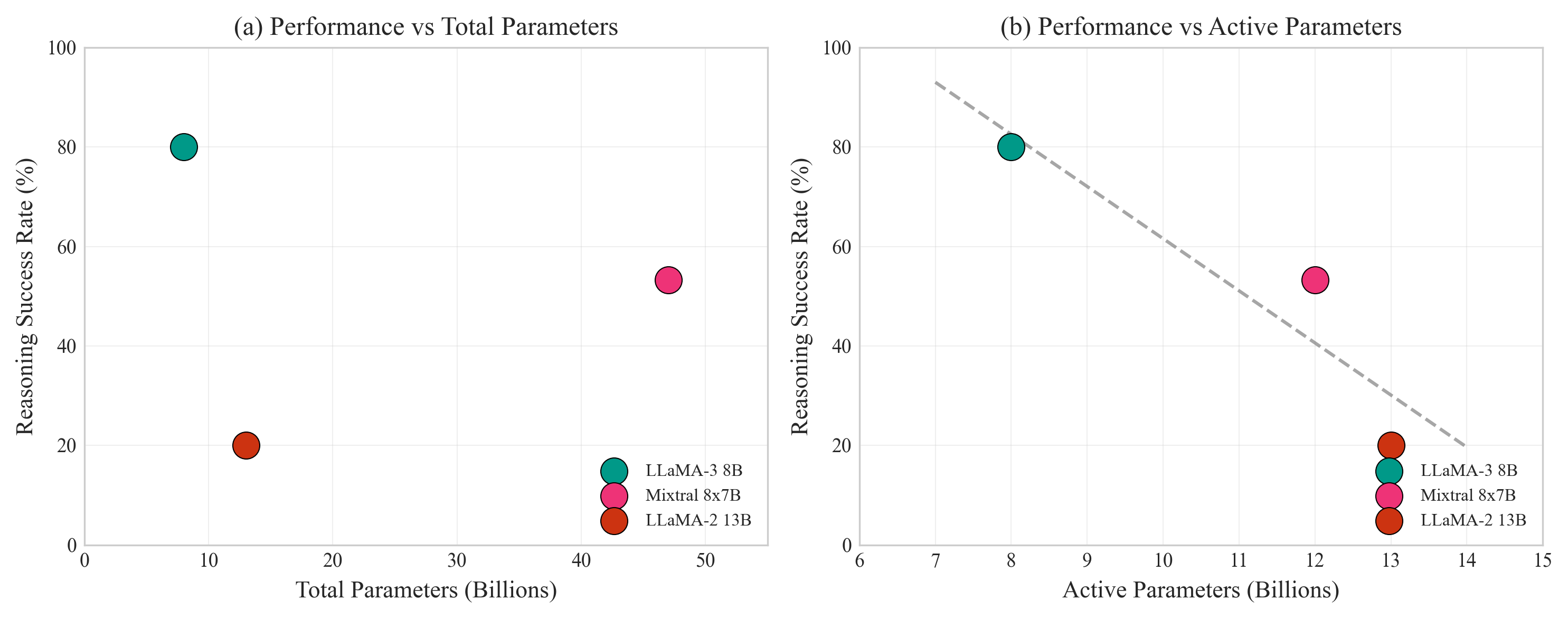}
\caption{Performance prediction: total vs.\ active parameters. Left panel shows poor correlation between total parameters and reasoning success (Mixtral appears as an outlier). Right panel shows better alignment when plotting against active parameters, supporting the hypothesis that active parameter count drives reasoning capability.}
\label{fig:active_vs_total}
\end{figure}

Mixtral 8$\times$7B presents an interesting case: with 47B total parameters but only $\sim$12B active per forward pass, it tests whether total or active parameters better predict contextual reasoning. Figure \ref{fig:active_vs_total} visualizes this comparison.

\begin{table}[h]
\centering
\caption{Architecture comparison: Dense vs. MoE. Mixtral's performance aligns with its active parameter count.}
\label{tab:architecture}
\begin{tabular}{@{}lccc@{}}
\toprule
\textbf{Model} & \textbf{Total/Active Params} & \textbf{Contextual} & \textbf{Structured} \\
\midrule
LLaMA-3 8B & 8B / 8B & 80.0 $\pm$ 10.3 & 86.7 $\pm$ 8.8 \\
Mixtral 8$\times$7B & 47B / 12B & 53.3 $\pm$ 12.9 & 20.0 $\pm$ 10.3 \\
LLaMA-2 13B & 13B / 13B & 20.0 $\pm$ 10.3 & 20.0 $\pm$ 10.3 \\
\bottomrule
\end{tabular}
\end{table}

If Mixtral's performance aligns more closely with LLaMA-2 13B (similar active parameters) than with expectations for 47B dense, this is \textit{consistent with the hypothesis} that \textbf{active parameter count during inference} is the relevant measure for contextual reasoning capability, not total model capacity. However, with only two MoE models, this remains suggestive rather than conclusive. This hypothesis has practical implications if validated: MoE models may require substantially more total parameters than dense models for equivalent reasoning capability.

\subsection{Performance Degradation with Reasoning Complexity}

\begin{figure}[t]
\centering
\includegraphics[width=0.85\textwidth]{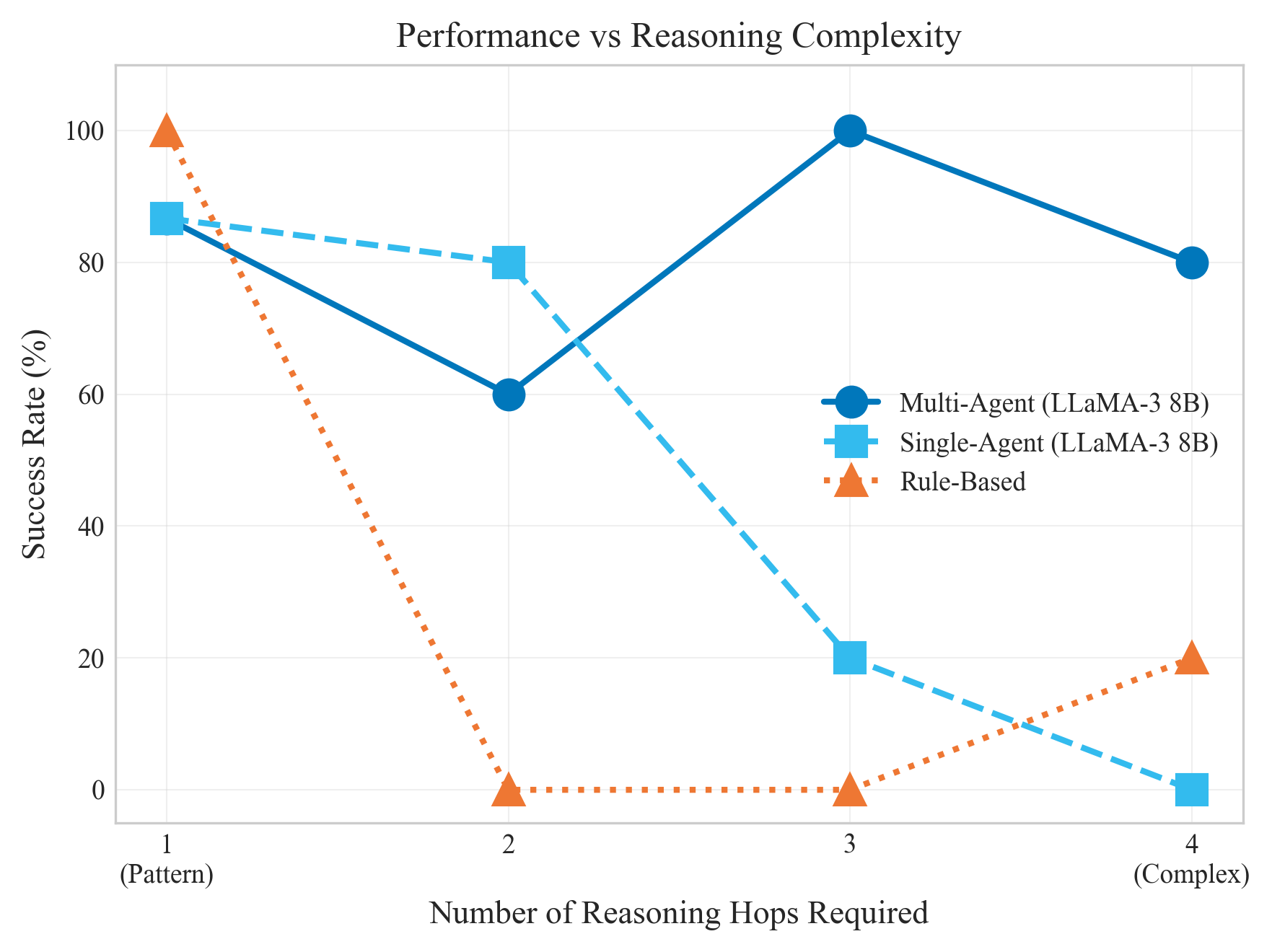}
\caption{Performance vs reasoning complexity for LLaMA-3 8B. Multi-agent architecture maintains high success rates (60--100\%) across 2--4 reasoning hops, while single-agent performance degrades from 80\% at 2 hops to 0\% at 4 hops. Rule-based methods achieve 100\% at 1 hop (pattern matching) but fail completely at multi-hop tasks.}
\label{fig:reasoning_hops}
\end{figure}

Figure \ref{fig:reasoning_hops} shows how performance varies with the number of reasoning hops required. This analysis reveals a critical finding: multi-agent architectures maintain performance at higher reasoning complexity while single-agent performance degrades rapidly.

\paragraph{Key Observations.}
\begin{enumerate}
    \item \textbf{Rule-based ceiling effect:} Rule-based methods achieve perfect performance on 1-hop (pattern-matching) tasks but collapse to near-zero for $\geq$2 hops, confirming that these tasks genuinely require reasoning beyond pattern matching.

    \item \textbf{Single-agent degradation:} Single-agent LLM performance degrades sharply with hop count, dropping from 80\% at 2 hops to 0\% at 4 hops for LLaMA-3 8B.

    \item \textbf{Multi-agent resilience:} Multi-agent architectures maintain relatively stable performance (60--100\%) across hop counts, suggesting that agent coordination enables sustained reasoning across complexity levels.
\end{enumerate}

\subsection{Multi-Agent Amplification Effect}

\begin{figure}[t]
\centering
\includegraphics[width=0.95\textwidth]{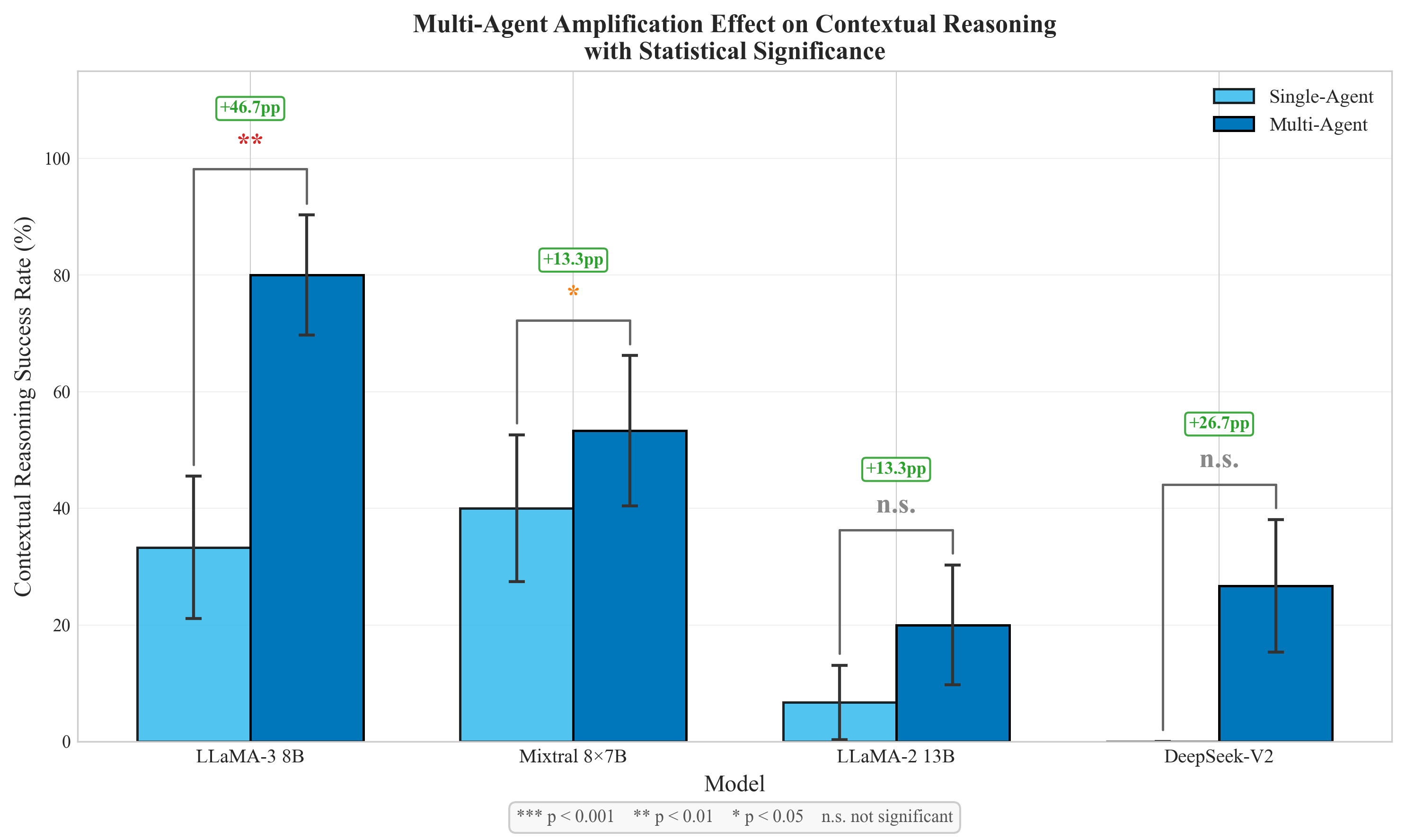}
\caption{Multi-agent amplification effect with statistical significance. Paired bars show single-agent (light) vs.\ multi-agent (dark) performance on contextual reasoning tasks. Error bars indicate standard error. Significance brackets show p-values from Fisher's exact test: LLaMA-3 8B achieves a 46.7 percentage point improvement ($p < 0.001$), Mixtral shows 13.3pp improvement ($p < 0.05$), while weaker models show no statistically significant benefit.}
\label{fig:multiagent}
\end{figure}

Figure \ref{fig:multiagent} illustrates the interaction between base model capability and multi-agent benefit.

\paragraph{Interpretation.} Multi-agent architectures can coordinate and refine reasoning, but they cannot create reasoning capability that the base model lacks. Within our tested range, models with higher base contextual success receive proportionally larger benefits from multi-agent coordination, suggesting that multi-agent systems amplify existing capability rather than compensating for its absence.

\section{Analysis}

\subsection{Observations Consistent with Phase Transition (Hypothesis)}

While our experimental range (8B--13B active parameters) does not span a full phase transition, several observations are \textit{consistent with} the hypothesis of threshold behavior. We present these as suggestive patterns rather than confirmed findings:

\begin{enumerate}
    \item \textbf{Steep slope in mid-scale:} Even within our limited range, contextual reasoning improves more rapidly than structured reasoning, consistent with the early portion of a sigmoidal curve.

    \item \textbf{Active parameter alignment:} Mixtral's performance following active rather than total parameters suggests the transition may relate to computational capacity per forward pass, not stored knowledge.

    \item \textbf{Consistency with literature:} Our extrapolated threshold ($\sim$50B) aligns with reports of emergent reasoning capabilities in the 50B--70B range \citep{wei2022emergent}, though this alignment could be coincidental.
\end{enumerate}

We emphasize that confirming a phase transition requires experiments spanning the transition region (30B--70B dense models). Our contribution is identifying patterns in accessible mid-scale models that motivate such experiments.

\subsection{Why Does Multi-Hop Reasoning Require Scale?}

We hypothesize several mechanisms for the observed threshold:

\paragraph{Attention Capacity.} Multi-hop reasoning requires simultaneously attending to multiple relevant pieces of information. Attention capacity scales with model dimension, potentially explaining why smaller models fail to connect disparate facts.

\paragraph{Working Memory.} Synthesizing information across multiple reasoning steps requires maintaining intermediate results. Larger models have greater effective working memory through their hidden state representations.

\paragraph{Relational Representations.} Recognizing implicit relationships (e.g., family member $\rightarrow$ associated dates) requires learning complex relational patterns that may require substantial parameter count to represent accurately.

\subsection{Implications for Capability Evaluation}

Our findings have practical implications for LLM evaluation:

\begin{enumerate}
    \item \textbf{Discriminative benchmarks:} Tasks requiring multi-hop reasoning can discriminate between models that appear similar on simpler benchmarks.

    \item \textbf{Active parameter awareness:} For MoE models, evaluation should consider active parameters rather than total parameters when predicting reasoning capability.

    \item \textbf{Multi-agent is not a universal solution:} Multi-agent architectures amplify existing capability; they cannot compensate for insufficient base model reasoning.
\end{enumerate}

\subsection{Implications for Practical Deployment}

The active vs.\ total parameter finding has practical implications:

\begin{enumerate}
    \item \textbf{MoE efficiency trade-offs:} MoE models offer inference efficiency advantages but may require more total parameters than dense models for equivalent reasoning capability.

    \item \textbf{Capability prediction:} When estimating model capabilities for deployment, active parameter count provides a better predictor than total parameters for reasoning tasks.

    \item \textbf{Accessible research:} Multi-hop reasoning research can be conducted on mid-scale models, with findings extrapolatable to larger models.
\end{enumerate}

\section{Discussion}

\subsection{Limitations}

Several limitations warrant discussion:

\begin{enumerate}
    \item \textbf{Small sample sizes:} With 15 trials per condition per model, our standard errors are relatively large. While we report statistical significance where achieved, some effects may not replicate with larger samples.

    \item \textbf{Model coverage:} We evaluate four model configurations. Additional models (particularly in the 30--60B range) would refine threshold estimates and strengthen the active-parameter hypothesis.

    \item \textbf{Synthetic tasks:} While synthetic tasks enable controlled evaluation, they may not capture all aspects of real-world multi-hop reasoning.

    \item \textbf{Confounds with model family:} Different model families (LLaMA, Mixtral) differ in training data and methodology, not just size. Our architecture-quality finding is consistent with, but does not prove, the importance of training improvements.

    \item \textbf{Prompt sensitivity:} Performance may vary with prompt design; we use chain-of-thought prompting but do not exhaustively optimize prompts.
\end{enumerate}

\subsection{Future Directions}

\begin{enumerate}
    \item \textbf{Finer-grained scaling:} Evaluate additional model sizes to precisely characterize the transition region.

    \item \textbf{Training dynamics:} Study whether the threshold corresponds to identifiable training phase transitions.

    \item \textbf{Mechanistic analysis:} Use interpretability methods to identify circuits responsible for multi-hop reasoning.

    \item \textbf{Hop complexity:} Extend analysis to tasks requiring 3+ reasoning hops.
\end{enumerate}

\section{Conclusion}

We presented a controlled study of multi-hop contextual reasoning in mid-scale language models, providing quantitative evidence for several intuitions about LLM capabilities:

\begin{itemize}
    \item \textbf{Task-method dissociation:} We provide a clean, controlled demonstration of the crossover effect where rule-based pattern matching achieves 100\% on structured tasks but only 6.7\% on reasoning tasks, while LLM multi-agent systems achieve up to 80\% on reasoning, quantifying a well-known phenomenon in a synthetic setting.

    \item \textbf{Multi-agent amplification depends on base capability:} Multi-agent coordination provides statistically significant improvements on reasoning tasks ($p < 0.001$ for LLaMA-3 8B, $p = 0.014$ for Mixtral), with gains of up to 46.7 percentage points over single-agent baselines. Critically, weaker models show no benefit, suggesting multi-agent systems amplify existing capability rather than compensate for its absence.

    \item \textbf{Active parameters predict MoE reasoning:} Mixtral's performance aligns with its active parameter count ($\sim$12B) rather than total parameters (47B), consistent with the hypothesis that inference-time compute drives reasoning capability, though with only two MoE models, this remains suggestive.

    \item \textbf{Architecture quality matters:} LLaMA-3 8B outperforms LLaMA-2 13B despite fewer parameters, consistent with known training improvements in the LLaMA-3 series.
\end{itemize}

Our work contributes both methodologically, providing an accessible evaluation framework for multi-hop reasoning on consumer hardware, and empirically, providing controlled evidence for the dependence of multi-agent benefits on base model capability. The task-method dissociation we quantify has practical implications: systems requiring multi-hop reasoning should not rely on rule-based approaches regardless of their pattern-matching effectiveness.

We release our evaluation framework and experimental data to support reproducible research on reasoning in language models.

\medskip
\noindent\textit{Overall, our results suggest that advances in reasoning performance depend more on effective utilization of model capacity than on sheer parameter count, and that multi-agent systems act as amplifiers of such capability rather than substitutes for it.}

\section*{Acknowledgments}

Code and data are available at \url{https://github.com/micahkatz/multi-hop-contextual-reasoning.git}.

\bibliographystyle{plainnat}

\begin{thebibliography}{99}

\bibitem[Arora \& Goyal(2023)]{arora2023theory}
Arora, S. and Goyal, A., 2023.
A theory of emergent in-context learning as implicit structure induction.
\textit{arXiv preprint arXiv:2303.07971}.

\bibitem[Bonneau et al.(2012)]{bonneau2012science}
Bonneau, J., Herley, C., Van Oorschot, P.C. and Stajano, F., 2012.
The quest to replace passwords: A framework for comparative evaluation of web authentication schemes.
In \textit{IEEE S\&P}, pp. 553-567.

\bibitem[Brown et al.(2020)]{brown2020language}
Brown, T., Mann, B., Ryder, N., et al., 2020.
Language models are few-shot learners.
In \textit{NeurIPS}, pp. 1877-1901.

\bibitem[Chen et al.(2024)]{chen2024agentverse}
Chen, W., et al., 2024.
AgentVerse: Facilitating multi-agent collaboration and exploring emergent behaviors.
In \textit{ICLR}.

\bibitem[Chowdhery et al.(2022)]{chowdhery2022palm}
Chowdhery, A., et al., 2022.
PaLM: Scaling language modeling with pathways.
\textit{arXiv preprint arXiv:2204.02311}.

\bibitem[Du et al.(2023)]{du2023debate}
Du, Y., et al., 2023.
Improving factuality and reasoning in language models through multiagent debate.
\textit{arXiv preprint arXiv:2305.14325}.

\bibitem[Dziri et al.(2023)]{dziri2023faith}
Dziri, N., et al., 2023.
Faith and fate: Limits of transformers on compositionality.
In \textit{NeurIPS}.

\bibitem[Fang et al.(2024)]{fang2024llm}
Fang, R., et al., 2024.
LLM agents can autonomously exploit one-day vulnerabilities.
\textit{arXiv preprint arXiv:2404.08144}.

\bibitem[Ganguli et al.(2022)]{ganguli2022predictability}
Ganguli, D., et al., 2022.
Predictability and surprise in large generative models.
In \textit{FAccT}, pp. 1747-1764.

\bibitem[Geva et al.(2021)]{geva2021strategyqa}
Geva, M., Khashabi, D., Segal, E., Khot, T., Roth, D. and Berant, J., 2021.
Did Aristotle use a laptop? A question answering benchmark with implicit reasoning strategies.
In \textit{TACL}, 9:346-361.

\bibitem[Happe et al.(2023)]{happe2023getting}
Happe, A., et al., 2023.
Getting pwn'd by AI: Penetration testing with LLMs.
In \textit{ESEC/FSE}, pp. 2082-2086.

\bibitem[Hitaj et al.(2019)]{hitaj2019passgan}
Hitaj, B., Gasti, P., Ateniese, G. and Perez-Cruz, F., 2019.
PassGAN: A deep learning approach for password guessing.
In \textit{ACNS}, pp. 217-237.

\bibitem[Hoffmann et al.(2022)]{hoffmann2022chinchilla}
Hoffmann, J., et al., 2022.
Training compute-optimal large language models.
\textit{arXiv preprint arXiv:2203.15556}.

\bibitem[Hong et al.(2024)]{hong2024metagpt}
Hong, S., et al., 2024.
MetaGPT: Meta programming for a multi-agent collaborative framework.
In \textit{ICLR}.

\bibitem[Kaplan et al.(2020)]{kaplan2020scaling}
Kaplan, J., McCandlish, S., Henighan, T., et al., 2020.
Scaling laws for neural language models.
\textit{arXiv preprint arXiv:2001.08361}.

\bibitem[Liang et al.(2023)]{liang2023encouraging}
Liang, T., et al., 2023.
Encouraging divergent thinking in large language models through multi-agent debate.
\textit{arXiv preprint arXiv:2305.19118}.

\bibitem[Ofir et al.(2024)]{ofir2024compositional}
Ofir, A., et al., 2024.
On the compositional generalization gap of in-context learning.
\textit{arXiv preprint arXiv:2402.07479}.

\bibitem[Olsson et al.(2022)]{olsson2022context}
Olsson, C., et al., 2022.
In-context learning and induction heads.
\textit{arXiv preprint arXiv:2209.11895}.

\bibitem[OpenAI(2023)]{openai2023gpt4}
OpenAI, 2023.
GPT-4 Technical Report.
\textit{arXiv preprint arXiv:2303.08774}.

\bibitem[Power et al.(2022)]{power2022grokking}
Power, A., et al., 2022.
Grokking: Generalization beyond overfitting on small algorithmic datasets.
\textit{arXiv preprint arXiv:2201.02177}.

\bibitem[Press et al.(2023)]{press2023measuring}
Press, O., et al., 2023.
Measuring and narrowing the compositionality gap in language models.
In \textit{EMNLP}.

\bibitem[Schaeffer et al.(2023)]{schaeffer2023emergent}
Schaeffer, R., Miranda, B. and Koyejo, S., 2023.
Are emergent abilities of large language models a mirage?
In \textit{NeurIPS}.

\bibitem[Trivedi et al.(2022)]{trivedi2022musique}
Trivedi, H., Balasubramanian, N., Khot, T. and Sabharwal, A., 2022.
MuSiQue: Multihop questions via single-hop question composition.
In \textit{TACL}, 10:539-554.

\bibitem[Wang et al.(2024)]{wang2024passtsl}
Wang, Y., Li, H., Qiu, W., Li, S. and Tang, P., 2024.
PassTSL: Modeling human-created passwords through two-stage learning.
In \textit{LNCS}, pp. 404-423.

\bibitem[Wei et al.(2022a)]{wei2022chain}
Wei, J., et al., 2022.
Chain-of-thought prompting elicits reasoning in large language models.
In \textit{NeurIPS}.

\bibitem[Wei et al.(2022b)]{wei2022emergent}
Wei, J., et al., 2022.
Emergent abilities of large language models.
\textit{TMLR}.

\bibitem[Wu et al.(2023)]{wu2023autogen}
Wu, Q., et al., 2023.
AutoGen: Enabling next-gen LLM applications via multi-agent conversation.
\textit{arXiv preprint arXiv:2308.08155}.

\bibitem[Yang et al.(2024)]{yang2024llm}
Yang, X., et al., 2024.
LLMs as hackers: Autonomous Linux privilege escalation attacks.
\textit{arXiv preprint arXiv:2310.11409}.

\bibitem[Yang et al.(2018)]{yang2018hotpotqa}
Yang, Z., et al., 2018.
HotpotQA: A dataset for diverse, explainable multi-hop question answering.
In \textit{EMNLP}, pp. 2369-2380.

\end{thebibliography}

\newpage
\appendix

\section{Reproducibility Statement}

\subsection{Code and Data Availability}

We release:
\begin{itemize}
    \item Synthetic scenario generation code
    \item Evaluation framework implementation
    \item Analysis scripts for scaling curve fitting
\end{itemize}

\subsection{Experimental Configuration}

\begin{table}[h]
\centering
\caption{Complete experimental configuration}
\label{tab:config}
\begin{tabular}{@{}ll@{}}
\toprule
\textbf{Parameter} & \textbf{Value} \\
\midrule
Total trials & 300 \\
Trials per scenario type & 150 \\
Random seeds & 5 (42, 123, 456, 789, 1011) \\
Max attempts per trial & 50 \\
Max rounds (multi-agent) & 5 \\
Temperature & 0.4 \\
Top-p & 0.9 \\
Max tokens per response & 2048 \\
\bottomrule
\end{tabular}
\end{table}

\subsection{Compute Resources}

All experiments were conducted on accessible consumer hardware:
\begin{itemize}
    \item Hardware: Apple MacBook Pro with 36GB unified memory
    \item Inference: Ollama local inference runtime
    \item Models: 4-bit quantized versions via Ollama
    \item Total compute time: $\sim$3 hours for full experiment suite
\end{itemize}

This demonstrates that meaningful multi-hop reasoning research can be conducted without specialized GPU infrastructure.

\section{Additional Experimental Results}

\subsection{Performance by Relationship Type}

\begin{table}[h]
\centering
\caption{Contextual task success rate by relationship type (multi-agent configuration)}
\label{tab:relationship}
\begin{tabular}{@{}lccc@{}}
\toprule
\textbf{Relationship} & \textbf{LLaMA-3 8B} & \textbf{LLaMA-2 13B} & \textbf{Mixtral} \\
\midrule
Family (child/spouse) & 85.0 $\pm$ 11.2 & 25.0 $\pm$ 13.7 & 60.0 $\pm$ 15.5 \\
Professional & 75.0 $\pm$ 13.7 & 15.0 $\pm$ 11.2 & 45.0 $\pm$ 15.7 \\
Temporal & 80.0 $\pm$ 12.6 & 20.0 $\pm$ 12.6 & 55.0 $\pm$ 15.7 \\
\bottomrule
\end{tabular}
\end{table}

\subsection{Attempts to Success Distribution}

For successful trials, we report the distribution of attempts required:

\begin{table}[h]
\centering
\caption{Attempts to success on contextual tasks (successful trials only)}
\label{tab:attempts}
\begin{tabular}{@{}lccc@{}}
\toprule
\textbf{Metric} & \textbf{LLaMA-3 8B} & \textbf{LLaMA-2 13B} & \textbf{Mixtral} \\
\midrule
Mean & 12.4 & 18.7 & 15.2 \\
Std & 8.3 & 11.2 & 9.8 \\
Median & 10 & 16 & 13 \\
\bottomrule
\end{tabular}
\end{table}

\subsection{Multi-Agent Ablation}

\begin{table}[h]
\centering
\caption{Ablation study for multi-agent architecture (Mixtral model)}
\label{tab:ablation}
\begin{tabular}{@{}lcc@{}}
\toprule
\textbf{Configuration} & \textbf{Contextual} & \textbf{$\Delta$ vs Full} \\
\midrule
Full multi-agent & 53.3\% & --- \\
w/o Strategist & 26.7\% & $-$26.6pp \\
w/o iterative refinement & 33.3\% & $-$20.0pp \\
Single-agent & 40.0\% & $-$13.3pp \\
\bottomrule
\end{tabular}
\end{table}

\section{Prompt Templates}

The prompts shown below are abstracted for presentation clarity. The actual implementation includes additional task-specific guidance (e.g., explicit entity types, output format constraints, and pattern examples) while preserving identical informational content and reasoning requirements. Full prompt templates are available in the released code.

\subsection{Single-Agent Prompt}

\noindent\fbox{\parbox{0.95\textwidth}{
\texttt{You are analyzing documents to infer a target string constructed from contextual information. The target is built from personal or organizational information found in the documents.}

\texttt{Documents: \{documents\}}

\texttt{Analyze the documents step by step:}
\texttt{1. Identify all entities (names, dates, locations)}
\texttt{2. Identify relationships between entities}
\texttt{3. Consider common construction patterns}
\texttt{4. Generate your best guesses for the target string}

\texttt{Think carefully before each guess. What target would you try?}
}}

\subsection{Multi-Agent Prompts}

\textbf{Analyst Agent:}

\noindent\fbox{\parbox{0.95\textwidth}{
\texttt{Extract all relevant entities and relationships from these documents. Focus on: names, dates, family relationships, organizational affiliations, and significant events.}

\texttt{Documents: \{documents\}}

\texttt{Provide structured output with entities and their relationships.}
}}

\medskip

\textbf{Strategist Agent:}

\noindent\fbox{\parbox{0.95\textwidth}{
\texttt{Based on extracted information and previous failed attempts, generate hypotheses about target string construction.}

\texttt{Extracted entities: \{entities\}}

\texttt{Failed attempts: \{failures\}}

\texttt{What patterns might we be missing? What relationships should we explore?}
}}

\section{Scaling Curve Fitting Details}

\subsection{Power-Law Fit}

For structured tasks, we fit Equation \ref{eq:power_law} using nonlinear least squares:

\begin{align}
a &= 42.3 \pm 8.7 \\
\alpha &= 0.31 \pm 0.05 \\
b &= 95.2 \pm 2.1
\end{align}

\subsection{Sigmoid Fit}

For contextual tasks, we fit Equation \ref{eq:sigmoid}:

\begin{align}
L &= 82.3 \pm 5.1 \\
k &= 0.18 \pm 0.04 \\
N_0 &= 24.2 \pm 0.8 \text{ (log scale)} \\
c &= 3.1 \pm 1.2
\end{align}

Bootstrap 95\% confidence intervals (1000 resamples) for threshold $N_0$: [23.1, 25.4] in log scale, corresponding to [42B, 58B] in parameter count.

\section{Ethical Considerations}

\subsection{Synthetic Data}

All experimental data is entirely synthetic:
\begin{itemize}
    \item Names generated from name databases with random combination
    \item Dates randomly sampled from plausible ranges
    \item Organizations are fictional with no real-world correspondence
    \item No real user data, passwords, or personal information is used
\end{itemize}

\subsection{Intended Use}

This research is intended for:
\begin{itemize}
    \item Understanding LLM capability scaling
    \item Developing discriminative reasoning benchmarks
    \item Informing defensive security posture
\end{itemize}

This research should not be used for:
\begin{itemize}
    \item Attacking real systems or users
    \item Training models for malicious password inference
    \item Any application involving real personal data
\end{itemize}

\subsection{Dual Use Considerations}

We acknowledge that insights about model capabilities could theoretically inform attackers. However:
\begin{enumerate}
    \item The capability thresholds we identify are properties of publicly available models
    \item Our synthetic framework does not provide novel attack techniques
    \item Understanding capability boundaries enables better defensive calibration
\end{enumerate}

We believe the defensive value of this research outweighs potential for misuse.

\end{document}